# Inaccuracy of an E-Dictionary and Its Influence on Chinese Language Users

Fanfei MENG, Xi WANG, Shiyang ZHANG, Lan LI

The Chinese University of Hong Kong, Shenzhen

120030001@link.cuhk.edu.cn; 120040088@link.cuhk.edu.cn; 120030104@link.cuhk.edu.cn; lanli@cuhk.edu.cn

**Abstract**

Electronic dictionaries have replaced most paper dictionaries and have become the primary learning tools in assisting L2 learners in expanding their lexical repertoire. Generally, users tend to assume the definitions in E-dictionaries are reliable and barely question the validity of the descriptions provided. The accuracy of major E-dictionaries is not commonly challenged, and the formation of online dictionary corpora is rarely discussed. Moreover, the study of dictionary use is rather limited, especially E-dictionary problems. This study takes a combined research approach of experiment, survey, and dictionary critique to investigate the most popular E-dictionary in China – Youdao. The experiment included retrospection and translation test with the purpose of finding out how E-dictionary can affect student learning. Over a hundred students in different disciplines were asked to translate sentences that included insufficiently defined words in Youdao dictionary. The dictionary consultation process was recorded to observe how some deficient definitions influence participants' understanding of the words. It is found that E-dictionaries with errors of incomplete and/or inaccurate definitions could lead to students' misinterpretation of the sentence. In addition, student consultation habits were found problematic. The causes of the deficient word definitions are also explained from a data processing point of view. For instance, translation mistakes are integrated by the application of Artificial Intelligence and machine learning technology in online dictionaries. It is suggested that the dictionary skills of users should be further trained, and improvement of online dictionaries should also be made with better AI technologies.

## 1. Introduction

The pervasive influence of electronic dictionaries on second-language learners cannot be underestimated. Online dictionaries have become vital instruments in language learning, their impacts extend beyond simply providing access to a word's fundamental meaning, context-specific definitions, and collocations. Their influence is transformative, shaping learners' receptive and productive vocabulary knowledge, including spelling, pronunciation, usage in diverse contexts, and collocation, all hinging on the user's dictionary skills (Nation, 1990).Youdao dictionary is one of the most popular online dictionaries in China, which has a cumulative stream of 22 million users by 2019 (GETChina Insights, 2019). This widespread adoption speaks to the key role E-dictionaries, particularly Youdao, play in shaping the lexical repertoire of language learners. However, while these tools are transformative, their efficacy can be compromised by inaccuracies or incomplete definitions, highlighting the harm of over-trusting in the validity of online dictionary.

Previous studies have examined how L2 learners' proficiency affects the impact of dictionary use on their search results, production quality, and perception. For example, dictionary tactics, habits, or L2 proficiency may influence the results of a learner's search, leading to changes in their perception or production abilities. Chon's (2009) research shows that using dictionaries to tackle lexical problems may result in secondary lexical problems within the dictionary entry due to poor L2 vocabulary proficiency and problematic dictionary skills. Boonmoh (2012) argues that participants' language skills and metacognitive knowledge affect the success of dictionary lookups by analyzing the word choices in their writing.

Other studies have shed light on issues with online dictionaries themselves, such as webpage design and technical challenges. As noted by Li and Xu (2005), learners' performance in dictionary consultation may depend partly on the dictionary itself. When searching for a word, learners' word choices may be influenced by the target word's forms, illustrative examples, the usage of italics, boldface, and colours, among other factors (Farina, Vrbinc, & Vrbinc, 2019). When learners are formulating their own written or spoken language content, such as essays, presentations, or discussions, they often encounter obstacles. Learners may struggle to find the right words due to issues such as incorrect spelling, a lack of appropriate translation, a lack of equivalent meanings, or a mismatch between the translation and the learner's preconceptions (Boonmoh, 2012). However, studies identifying specific faults in online dictionaries and examining whether these flaws influence learners' vocabulary development are generally lacking.

One of the primary criticisms of E-dictionary among users, as revealed by a study conducted by Müller-Spitzer and Koplenig (2011), is whether online dictionaries can provide users with accurate and dependable information. Neglecting content quality could lead to the loss of the lexicon's legitimacy, therefore it is highly necessary to rigorously inspect E-dictionaries' content reliability. Regrettably, research in this area is scarce. Past research has mainly focused on comparing and ranking various online dictionaries to determine their reliability and effectiveness. However, research on comparing digital dictionaries with conventional authoritative dictionaries are not common. This study aims to identify common errors and deficiencies in the most popular electronic dictionary among Chinese university students and explore how these issues may affect undergraduate students' vocabulary development in second language. The main purpose of this paper is to verify two hypotheses:

H1: There is a prevalent tendency among Chinese students to heavily rely on electronic dictionaries, with minimal scrutiny of the accuracy or appropriateness of the provided results.

H2: The potential inaccuracies present in online dictionaries may lead students to generate erroneous understandings in their second language.

## 2. Methodology

### 2.1. Participants

The research consisted of two parts, including a short survey and an experiment. Participants in this study are 105 undergraduates aged between 18 and 22 studying in a variety of majors. Their native language is Mandarin Chinese and they have learned English as a foreign language for more than 12 years. More than 97 percent of them scored above 130 out of 150 in China's NMET (National Matriculation English Test), indicating they have fairly high English proficiency.

### 2.2. Concise dictionary & authoritative dictionary in Youdao

Youdao is one of the most popular E-dictionary in China (Liu et al., 2020). It offers various functions and interfaces, including "dictionary", "translating", and "paper searching", with the "dictionary" function being the primary focus of this study.

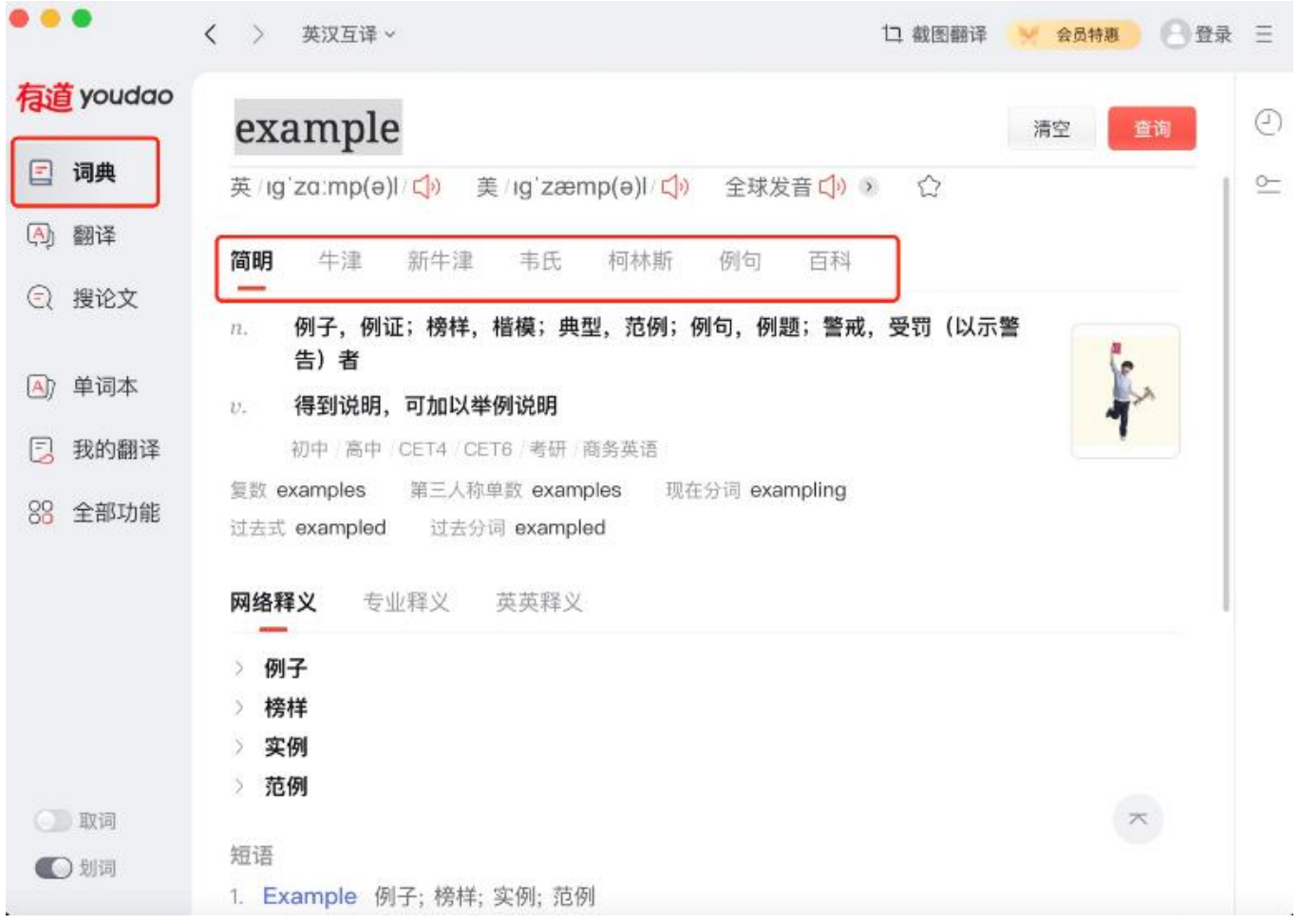

Figure 1: The general interface of Youdao Dictionary

The experiment in this study focused on analyzing word definition errors in the concise dictionary (CD), known as"Jian Ming" on the Youdao platform. To be more specific, the "dictionary" function provides users with five optional dictionaries to consult word definitions, including the concise dictionary (CD), *Oxford Dictionary*, *New Oxford American Dictionary*, *Webster's Dictionary*, and *Collins Dictionary* (see Figure 2). The word definitions in the CD are generated by the Youdao platform itself from extensive Internet resources. The last four dictionaries are all electronic copies of authoritative dictionaries (ADs) embedded into the Youdao platform. Our initial observation discovered that some word definitions in the Youdao concise dictionary are unreliable, despite it being the default page that users see first when they access the platform. Consequently, this study specifically targeted the concise dictionary.

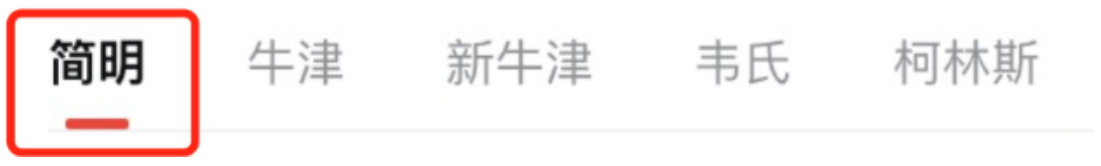

Figure 2: Five optional dictionaries under the "dictionary" function

### 2.3. Questionnaire design

To gain a comprehensive understanding of undergraduate students' usage of Youdao Dictionary, a survey questionnaire consisting of three sections was developed: participants' background information, their usage habits when using Youdao, and their overall attitudes towards Youdao (see Appendix 2). The questionnaire is presented in Chinese as participants are more comfortable with their native language and could complete the survey more quickly. The collected data was translated into English and used for subsequent cross-over analysis.

### 2.4. Translation experiment

In order to examine whether the word definition errors in the CD influence students' understanding of vocabulary, ten words with problematic Chinese definitions were selected for investigation. These words were provided by students who found them difficult in their English study activities, such as reading and writing essays, browsing websites, and watching films and television shows. The frequencies of the words were checked in two mega corpora: The Corpus of Contemporary American English (COCA) and The British National Corpus (BNC). The ten target words finally decided are: *furious*, *vigorous*, *dearly*, *fetish*, *raunchy*, *irritant*, *dicky*, *slosh*, *nostrum*, and *nerveless*. These words are unfamiliar words with frequencies lower than 1000 in COCA and BNC. The aim of word selection was to minimize the likelihood that students already know the word meanings during their dictionary consultation process.

Translation experiments were to investigate students' comprehension of new words. This approach has been widely recognized by researchers as an effective method for examining vocabulary learning. According to Nation (2013), unlike other forms of tests such as multiple-choice questions with interference options, translation tests require a more comprehensive grasp of vocabulary knowledge. It has also been demonstrated that learners in translational experiment focus more on vocabulary learning and use through the novel, and original reproduction of words than is the case in open composition tasks (Fageeh & Mekheimer, 2011). To ensure a clear assessment of students' translation abilities without interference, ten sentences hiding the selected ten tested words respectively have been crafted with simple syntactic structures and word expressions (see Appendix 1).

## 2.5. Research procedure

Both the survey and the experiment were conducted online, and all participants were required to share their screens with the researchers via Zoom for one-on-one observation. The questionnaire was set prior to the translation experiments with the aim of understanding students' attitudes towards Youdao and confirming the significance of studying the influence of dictionary accuracy on their learning.

Students were required to translate each sentence twice. In the first translation attempt, they referred to the concise dictionary (CD) only when encountering lexical problems. During the second-round translation, authoritative dictionaries (ADs) hyperlinked in Youdao, specifically *Oxford Dictionary*, *Webster's Dictionary*, and *Collins Dictionary*, could be used. The purpose of this comparison was to examine whether the definitions in two types of dictionaries would result in different understandings. There was no time limit for participants to complete the translation task, and the processes were monitored and screen-recorded for further analysis.

In addition, after completing the translation of each sentence, participants were asked two introspective questions. The first question elicited participants' instant feedback on how they had determined the meaning of the target word (Li & Xu, 2015), asking them whether their understanding of the tested word was based on the CD, the ADs, the context, or their prior knowledge. In the second introspective question, students were provided with the correct answers they had just worked on. By comparing the accurate answers with their own, they expressed their opinion on the quality of CD entries. Finally, positioned at the conclusion of the translation experiment, there was a final question for the participants that served as a comprehensive evaluation of the Concise Dictionary.

## 2.6. Data analysis

105 questionnaires were completed, focusing on 12 factors and their correlation with the participants' performance in the translation experiment with E-dictionaries. The factors included frequency usage, listening difficulty, speaking difficulty, reading difficulty, writing difficulty, translation difficulty, listening reading, writing translating, satisfaction level, vocabulary aid, reading aid, and word look-up (see Appendix 2).

Each participant' s translation was graded on a scale of 0 to 2, with 2 representing an accurate translation, 1 partially correct answer and 0 a wrong answer. Chi-square tests were conducted to investigate correlations between background factors and the total translation scores. Notably, significant correlations ($p<0.05$) were found in two factors: reading difficulty and vocabulary aid.

Descriptive statistical analysis was subsequently conducted with SPSS to delve deeper into these two factors. This included aggregating sub-categories of the factors against the TotalScore metric. Utilizing SPSS software, data was split into sub-items to better understand the influence of each factor on the final outcome.

In addition, a categorical summary analysis was performed, categorizing participants based on their sources of understanding in the translation. The sources included context, previous knowledge, the CD, and the ADs. The analysis shows that the highest score is by those who relied on previous word knowledge (84.6%) and the use of authoritative dictionaries (95%).

# 3. Results and Discussion

## 3.1. Picture of dictionary use

The questionnaire gained some insights into student dictionary use and their perception of Youdao's quality. Figure 3 reveals that Youdao is a highly popular dictionary, with about half of the participants using it multiple times a day, and 13% using it at least once a day. Moreover, among the five dictionaries embeded in Youdao, the concise dictionary is the most widely used, with 70% of users.

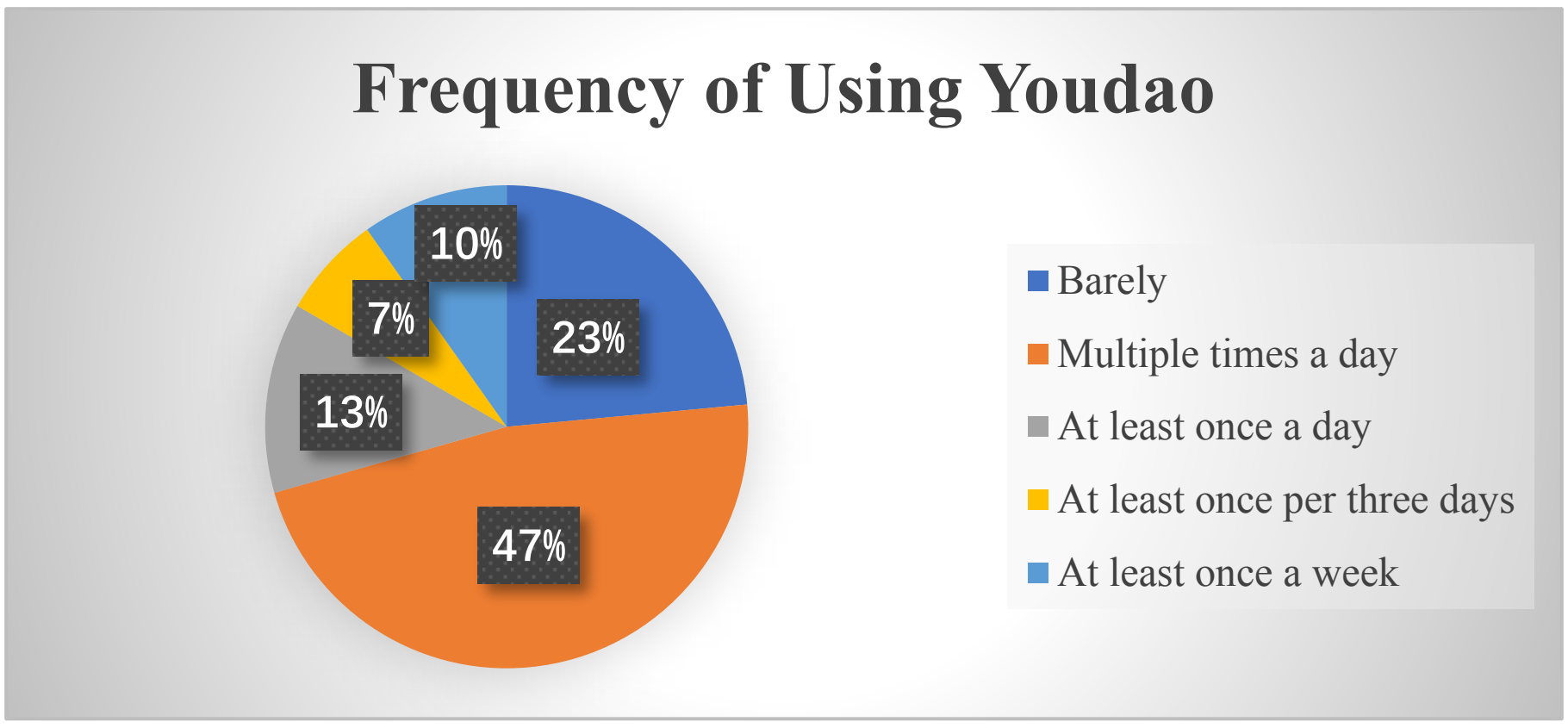

Figure 3. Frequency of using Youdao

The study also shows situations of dictionary use(Figure 4). 97% of participants consult dictionaries when encountering problems in reading, making it the most popular use scenario, followed by writing and translation experiment.

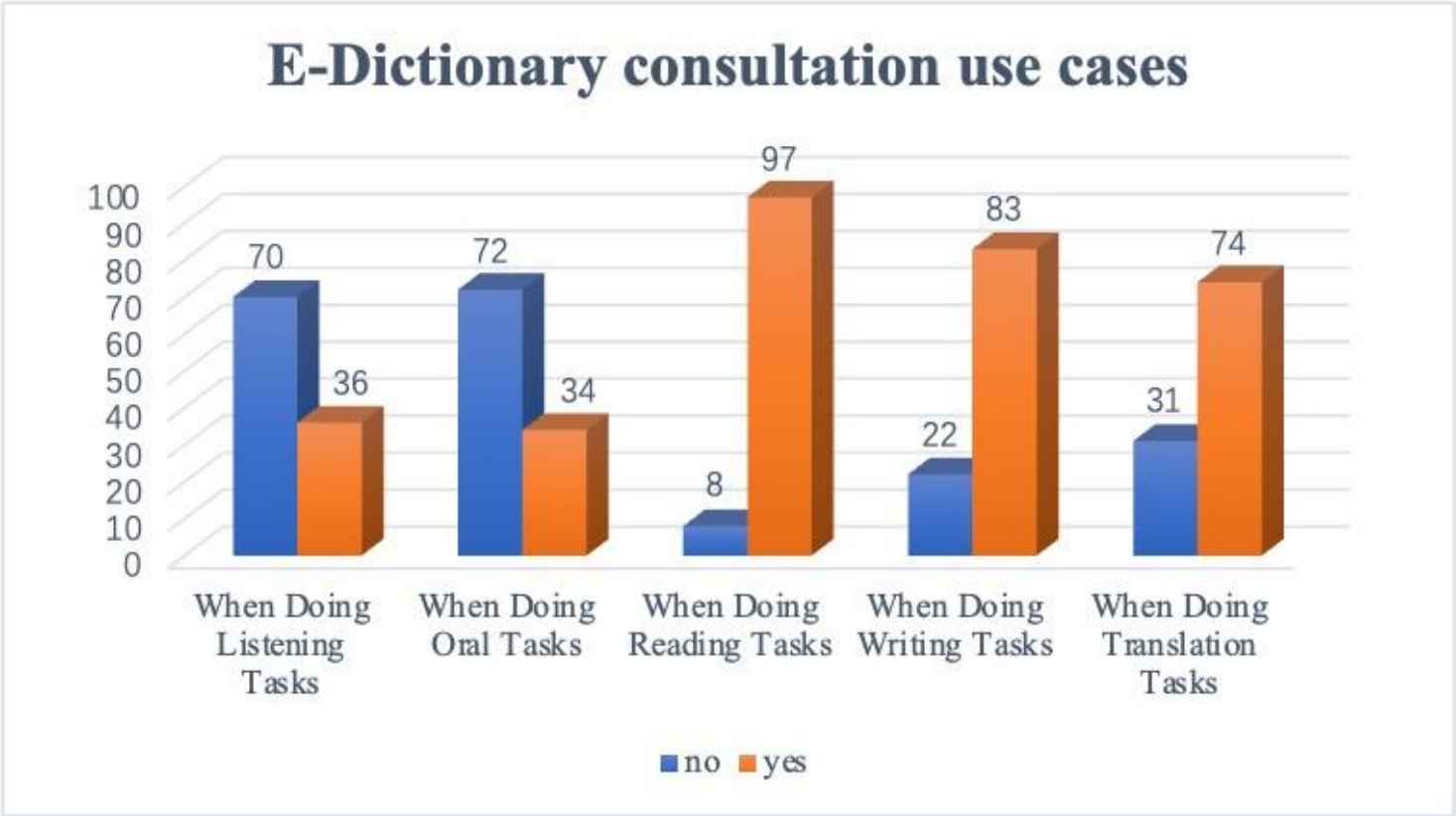

Figure 4. E-dictionary consultation use case

Interestingly, even with the awareness of the quality of the ADs, 69% of the participants preferred using the CD for convenience (Figure 5). Despite the considerable number of users depending on Youdao Concise Dictionary, the results indicate that users were not entirely satisfied with it. Figure 6 shows that only 8% of the participants deemed the definitions in the CD were very satisfactory; 36% were satisfactory and a significant portion, 43%, adopted a neutral position, and 12% thought they were very unsatisfactory. It can be seen that although Youdao CD is widely used by students, the quality of its definitions cannot keep up with users' expectations.

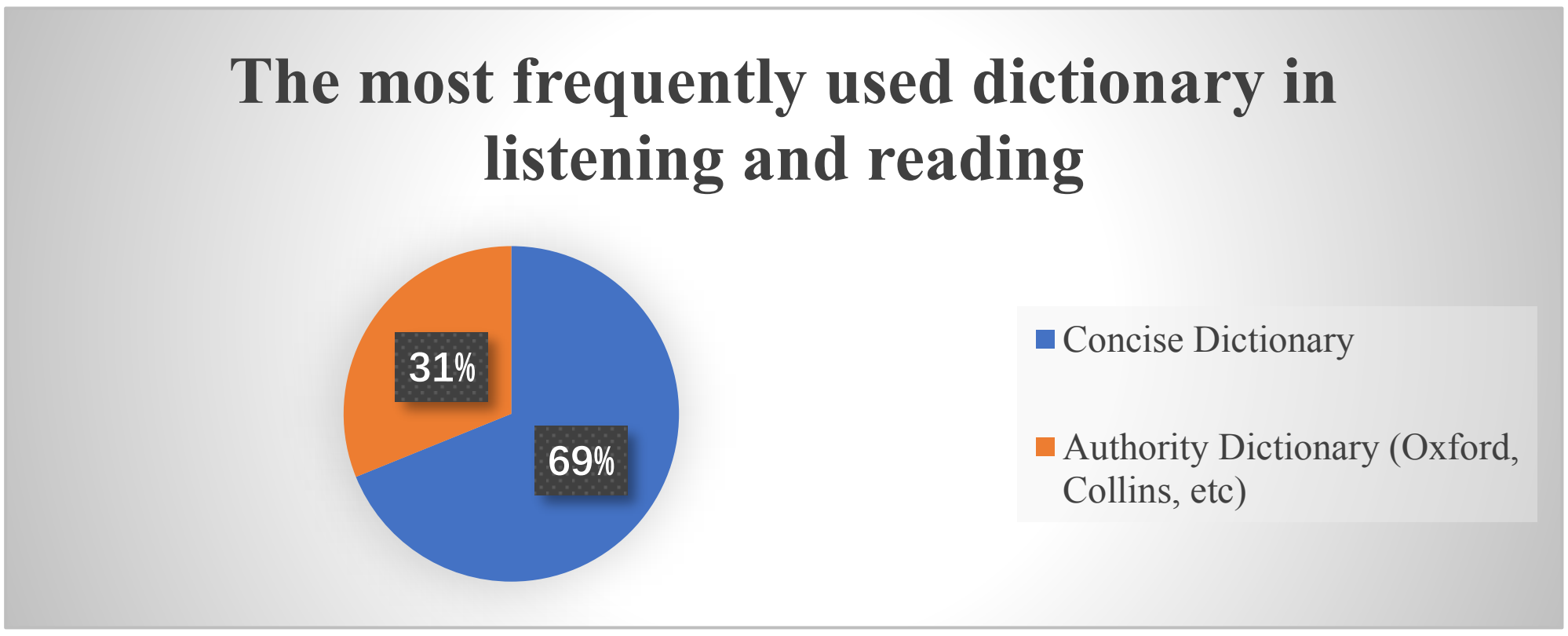

Figure 5. The most frequently used dictionary in listening and reading

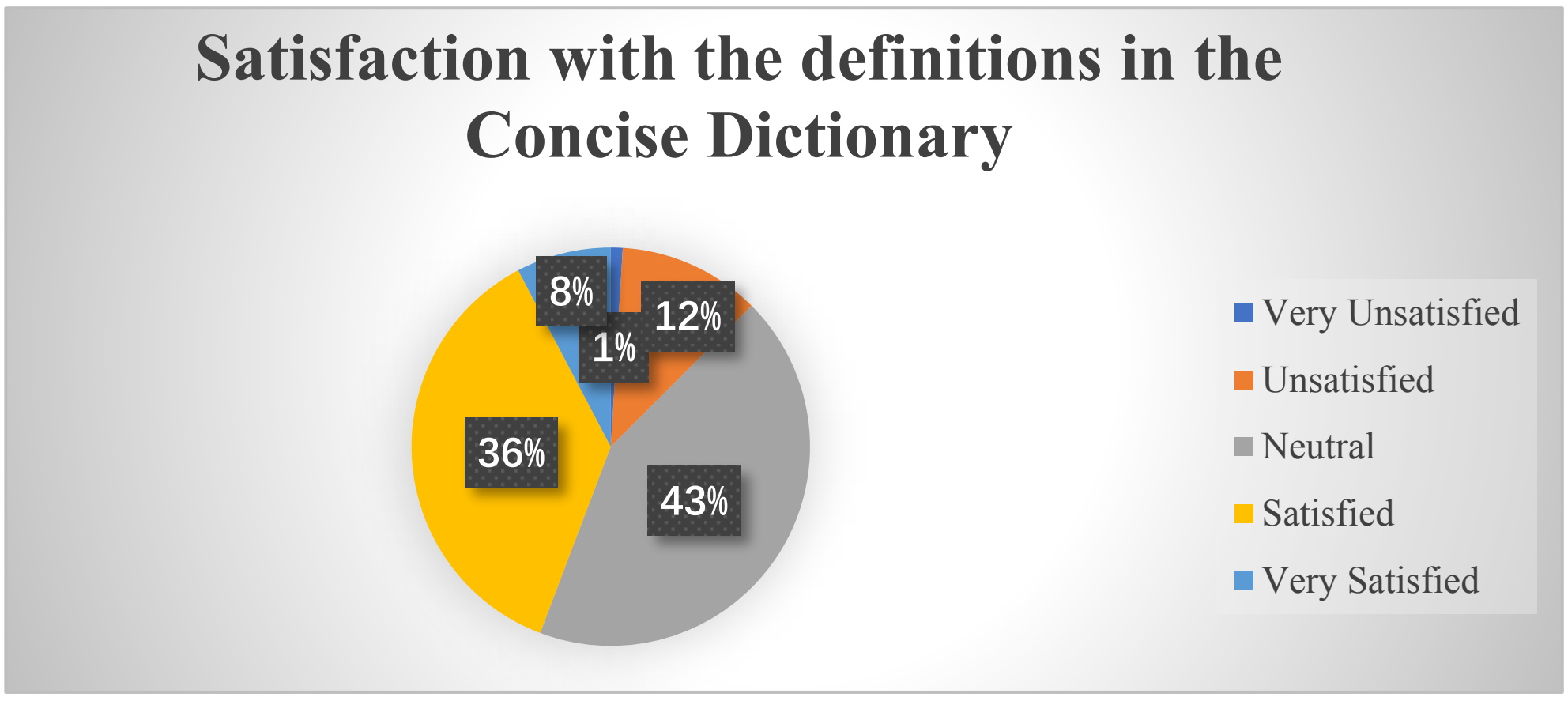

Figure 6. Satisfaction with the definitions in the Concise Dictionary

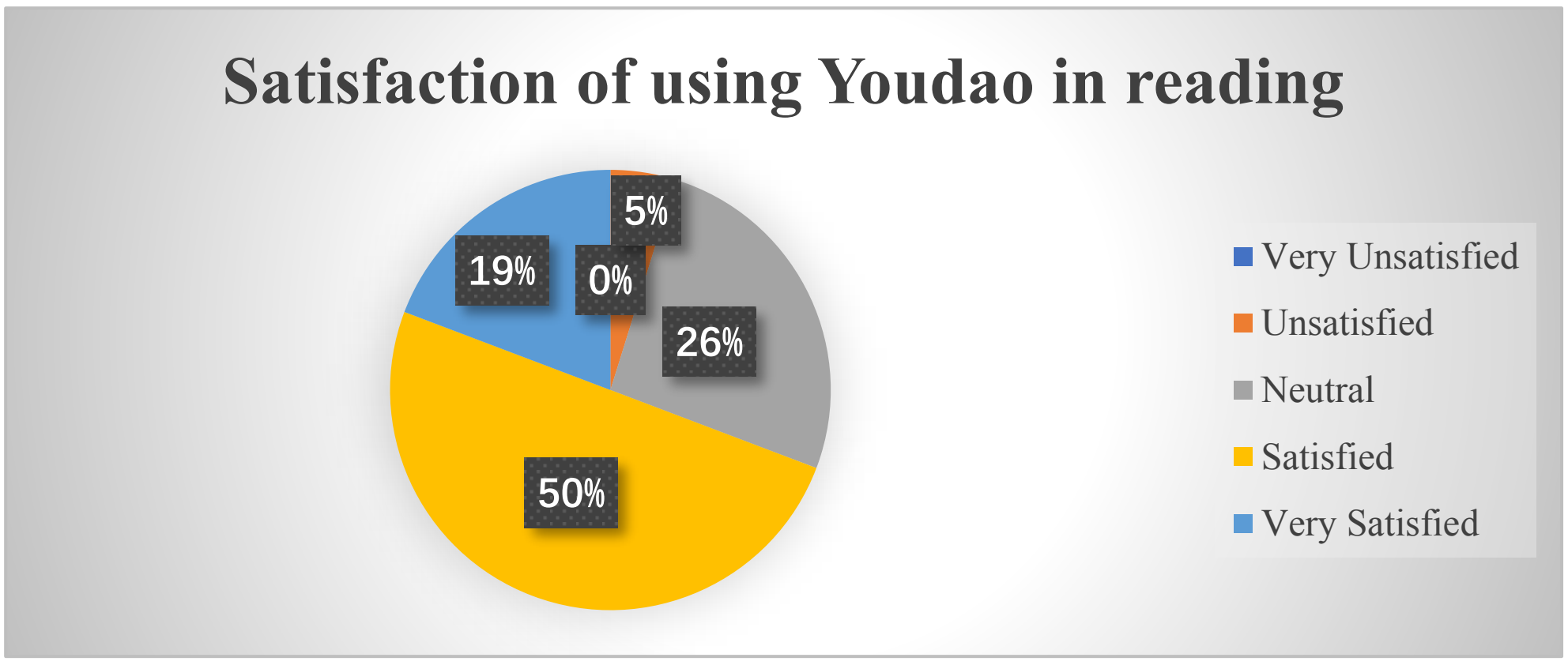

Figure 7. Satisfaction of using Youdao in reading

## 4. Analysis of dictionary use

### 4.1. Chi-square test (cross-analysis)

Since the questions were set in a 1-5 Leiker scale, they could be digitalized into statistical analysis. Each background question indicates an individual factor. In total, 12 different factors were set up: Frequency of Usage, Listening difficulty, Speaking Difficulty, Reading Difficulty, Writing Difficulty, Translation Difficulty, Listening, Reading, Writing, Translating, Satisfaction Level, Vocabulary Aid, Reading Aid, Word Look Up. Each factor and its corresponding question of the questionnaire can be found in Appendix 2. For example, Frequency of Usage is a factor that measures how often a participant uses online dictionary, where 1 indicates "Never" and 5 indicates "Everyday."

With the total score obtained from all the translation questions, Chi-square analysis was applied to each factor to examine if there is a clear correlation between any of the factors and the total performance. In other words, Categorical summary analysis investigates if any specific user behaviour can significantly affect their translating performance with the assistance of online dictionaries.

| Behavior Factor | $\chi^2$ | $\rho$ |
|---|---|---|
| FrequencyUsage | 48.005 | 0.944 |
| ListeningDificulty | 12.49 | 0.590 |
| SpeakingDifficulty | 8.483 | 0.811 |
| ReadingDitficulty | 38.459 | 0.000001 |
| WritingDificulty | 15.011 | 0.307 |
| TranslationDificulty | 18.447 | 0.141 |
| ListeningReading | 7.394 | 0.881 |
| WritingTranslating | 12.124 | 0.517 |
| SatisfactionLevel | 56.305 | 0.317 |
| VocabularyAid | 58.128 | 0.025 |
| ReadingAid | 51.593 | 0.085 |
| WordLsookUp | 44.952 | 0.745 |

Table 1. Chi-square analysis between hehavior factor and total score

Based on the table above, the use of chi-square test (cross-analysis) was conducted to examine the relationship between total score and the 12 different factors. It could be observed from the table that different Total Score samples did not exhibit significant differences ($p>0.05$) in 10 factors, except Reading Difficulty and Vocabulary Aid. Furthermore, analysis of Total Score samples demonstrated statistically significant variations ($p<0.05$) in two factors. This implied the presence of a statistical association between these factors and the Total Score. Consequently, participant behavior as measured by these two factors might exert some impacts on their translation performance.

## 4.2. Descriptive statistical analysis

While conducting a descriptive statistical analysis, we segmented individual factors into various sub-categories and then aggregated against the TotalScore metric to to gain insights into how each statistical behavior can impact the overall translation performance when using E-dictionaries.

Utilizing the SPSS software to identify patterns and trends in the data allowed for a more in-depth and accurate analysis of the data, which could be used to draw more meaningful conclusions and make more informed decisions. It is important to note that descriptive analysis is only one part of a broader analysis that includes other statistical techniques.

| Descriptive statistical analysis | | | | | |
|---|---|---|---|---|---|
| Question | Is Youdao dictionary helpful in learning new vocabulary? | | | | Summary |
| Options | Never | Barely | Sometimes | Always | |
| Number | 4 | 11 | 66 | 19 | 100 |
| Average | 10.25 | 10.091 | 10.242 | 11.158 | 10.4 |
| SD | 5.909 | 2.468 | 2.373 | 2.218 | 2.531 |

Table 2. Descriptive statistical analysis

## 4.3. Categorical summary analysis

Each participant in the study was required to translate sentences and indicated whether their understanding of the translated sentence came from the context, previous known, the concise dictionary or the authoritative dictionary. The participants' responses were then evaluated and classified into three categories – wrong (0 mark), partially correct (1 mark) and fully correct (2 marks). Using various benchmarks, the collected data was analyzed with the aid of Python programs.

To establish the statistical significance of the results, the data were further categorized into groups, such as those who claimed their understanding came from the context but got a score of 0, and those who used the CD and received full marks. The percentage composition of each category was then summarized and analyzed.

| **Category** | **irritant** | **nostrum** | **raunchy** | **nerveless** | **slosh** |
|---|---|---|---|---|---|
| CD_false（0） | 87% | 97% | 32% | 55% | 60% |
| CD_partially correct（1） | 3% | 3% | 61% | 9% | 30% |
| CD_correct（2） | 10% | 0% | 7% | 36% | 10% |
| ADs_ false（0） | 3% | 7% | 49% | 5% | 8% |
| ADs_ partially correct（1） | 12% | 9% | 8% | 0% | 23% |
| ADs_ correct（2） | 85% | 84% | 43% | 95% | 69% |

Table 3. Categorical summary analysis

## 5. Overview of the translation performance

In general, participants using the CD performed poorly. Under a grading standard that a score of 0 indicates a wrong answer, 1 represented a partially correct answer, and 2 indicated a correct answer, the average scores of 6 out of 10 tested words were below 1. Meanwhile, two scored between 1 to 1.5, and two scored higher than 1.5. It is important to note that the average scores for all tested words increased after consulting the authoritative dictionaries. The differences between consulting the two dictionaries ranged from 0.28 to 1.08.

| *Tested words* | *Score using the CD* | *Score using the Ads* | *Difference* |
|---|---|---|---|
| *dearly* | 0.82 | 1.06 | 0.24 |
| *furious* | 1.74 | 1.86 | 0.12 |
| *irritant* | 0.2 | 1.28 | 1.08 |
| *vigorous* | 1.2 | 1.6 | 0.4 |
| *slosh* | 0.4 | 1.04 | 0.64 |
| *raunchy* | 0.7 | 1.58 | 0.88 |
| *nostrum* | 0.08 | 0.82 | 0.74 |
| *dicky* | 1.22 | 1.6 | 0.38 |
| *fetish* | 1.5 | 1.78 | 0.28 |
| *nerveless* | 0.84 | 1.88 | 1.04 |

Table 4. Average point of the translation performance obtained through using two types of dictionaries

## 6. Analysis of the translation performance

As shown in Table 4, five words, *irritant*, *slosh*, *raunchy*, *nostrum*, and *nerveless* received the average score differences above 0.5. In the Categorical summary analysis (Table 3), the percentages of participants who consulted the CD but got 0 points in these items are 87%, 97%, 32%, 55% and 60% respectively. In order to explore the factors resulting in users' underperformance when using the CD, their translation and the interface of looking up the tested words in the CD were examined. Four factors influenced participants' understanding were summarized. Since *raunchy* and *irritant* belong to the same type of factor, only the latter one will be analyzed.

### 6.1. Deficient definition

The disparity between the scores obtained by using the CD and ADs to aid in the comprehension of *irritant* is remarkable, with a significant difference of 1.08. According to the categorical summary analysis(Table 3), only 10% of participants successfully captured the meaning of *irritant* within the given context and achieved full marks, while 87% of participants misunderstood the word. The source text and reference answer are presented below:

*He admitted that this problem was a main irritant.*

他承认这个问题是**让他恼火的**主要**原因**。

ta cheng ren zhe ge wen ti shi **rang ta nao huo de** zhu yao **yuan yin**.

Upon examining the definitions of *irritant* in both dictionaries within Youdao, a significant difference becomes evident. The CD provides a limited definition of 刺激物(ciji wu, sth irritating) and 刺激剂(ci ji ji, some solution irritating), aligning with the first sense presented in the *Oxford Dictionary* as "a substance that makes part of your body sore." However, it does not include the second sense in the *Oxford Dictionary*, "something that makes you annoyed or causes trouble."

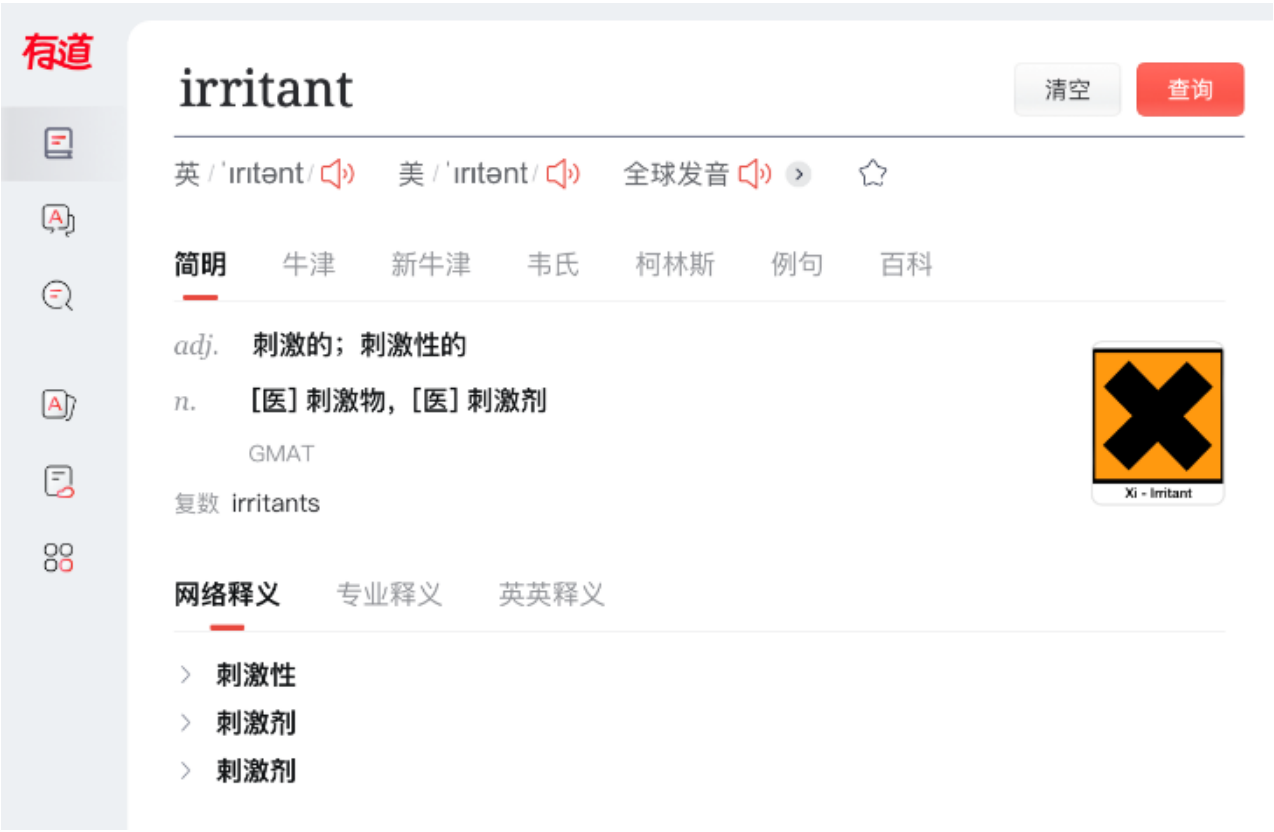

Figure 8. The entry of ***irritant*** in the concise dictionary

Consequently, if participants did not know the word before, it would be unlikely for them to understand the sentence accurately because they can only find one sense in the CD. Furthermore, they can not infer the meaning from the existing definition, as a substance that causes bodily soreness is unrelated to something that causes emotion. In the experiment, 63% of the translations were misled by this deficient definition. The students relied solely on the only definition of irritant in the CD, and produced wrong answers. For example：

他承认这个问题是一个主要的刺激物。

(ta cheng ren zhe ge wen ti shi yi ge zhu yao de ci ji wu.)

(He admitted that this problem was a main thing that stimulated him.)

## 6.2. Incorrect definition

Several definitions in the CD are erroneous or even contradictory to the correct meaning. A case in point is the word *nostrum*:

*Although my sister is not a doctor, she thinks she can cure any illness and is quick to suggest a nostrum to her friends.*

虽然我姐姐不是医生，但她认为她可以治愈任何疾病，并且很快就向她的朋友推荐了一种**江湖药**。

sui ran wo jie jie bu shi yi sheng ， dan ta ren wei ta ke yi zhi yu ren he ji bing ， bing qie hen kuai jiu xiang ta de peng you tui jian le yi zhong **jiang hu yao**.

Most participants referred to the CD and completely misunderstood the sentence with the word *nostrum.* The average score of the sentence was as low as 0.08, ranking the last among the 10 tested words. According to the *Oxford Dictionary*, *nostrum* is ideas or medicines that are often unscientific or ineffective; it has two senses：

1. (formal disapproving) an idea that is intended to solve a problem but that will probably not succeed 并非灵验的招数；不会奏效的计策
2. (old-fashioned) a medicine that is not made in a scientific way, and that is not effective 江湖药

However, the definitions in the CD suggest that *nostrum* means 秘方(mi fang, an undisclosed prescription with a marked medical effect) and 万能药 (wan neng yao, panacea), which are completely opposite to the original English definition: fraudulent medicine used to deceive people.

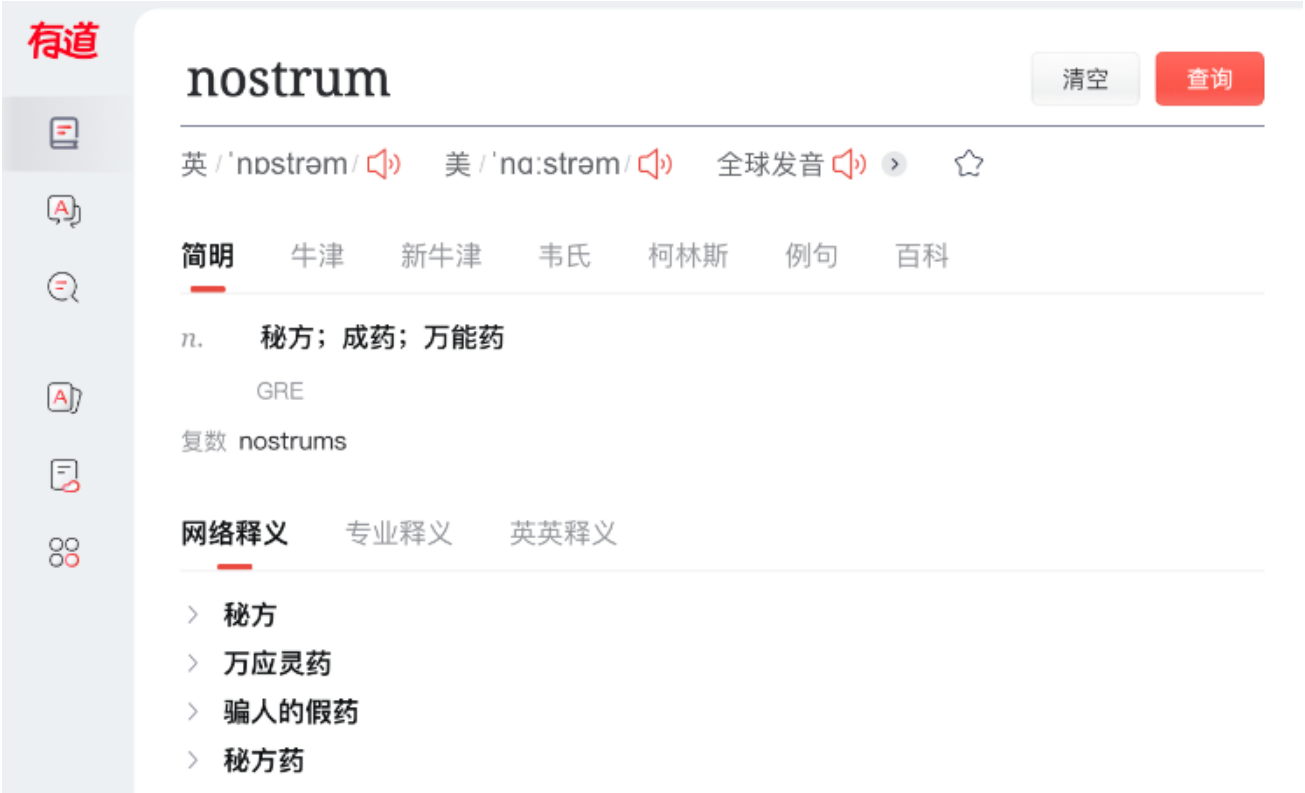

Figure 9: Screenshot of the concise dictionary’s interface of *nostrum*

Nevertheless, affected by the incorrect definition in the CD, 42% of the participants adopted 秘方(mi fang) to render this sentence, 17% of them understood *nostrum* as 灵丹妙药(ling dan miao yao)，万能药(wan neng yao)，很灵的药(hen ling de yao) , etc., all indicate the medicine’s effectiveness. In the introspective questionnaire, some participants pointed out the problems in *nostrum*’s definition:

1. The definitions in the CD is contradictory with each other.
2. Its definition is confusing and not accurate.
3. The definition is not that correct, which is easy to cause misunderstanding.

## 6.3. Incorrect and deficient definition

Another word caused low score in the sentence was the adjective *nerveless*. Those who used CD obtained an average score 0.84, while those consulted the ADs scored 1.88, showing a very striking difference. The Categorical summary analysis further unveiled noteworthy outcomes. When exclusively referring to the Concise Dictionary, 54% of the users misunderstood the sentence meaning. The source text and reference answer are as follows:

*He is one of the most nerveless champions in football history.*

他是足球历史上最**无畏的**冠军之一。

ta shi zu qiu li shi shang zui **wu wei de** guan jun zhi yi.

The issue stemmed from the CD's inadequate definitions. The second sense of the word, 'having no fear', is absent in the CD. Furthermore, the CD provides a completely opposite meaning of *nerveless* as 无勇气的 (wu yongqi de, lacking courage).

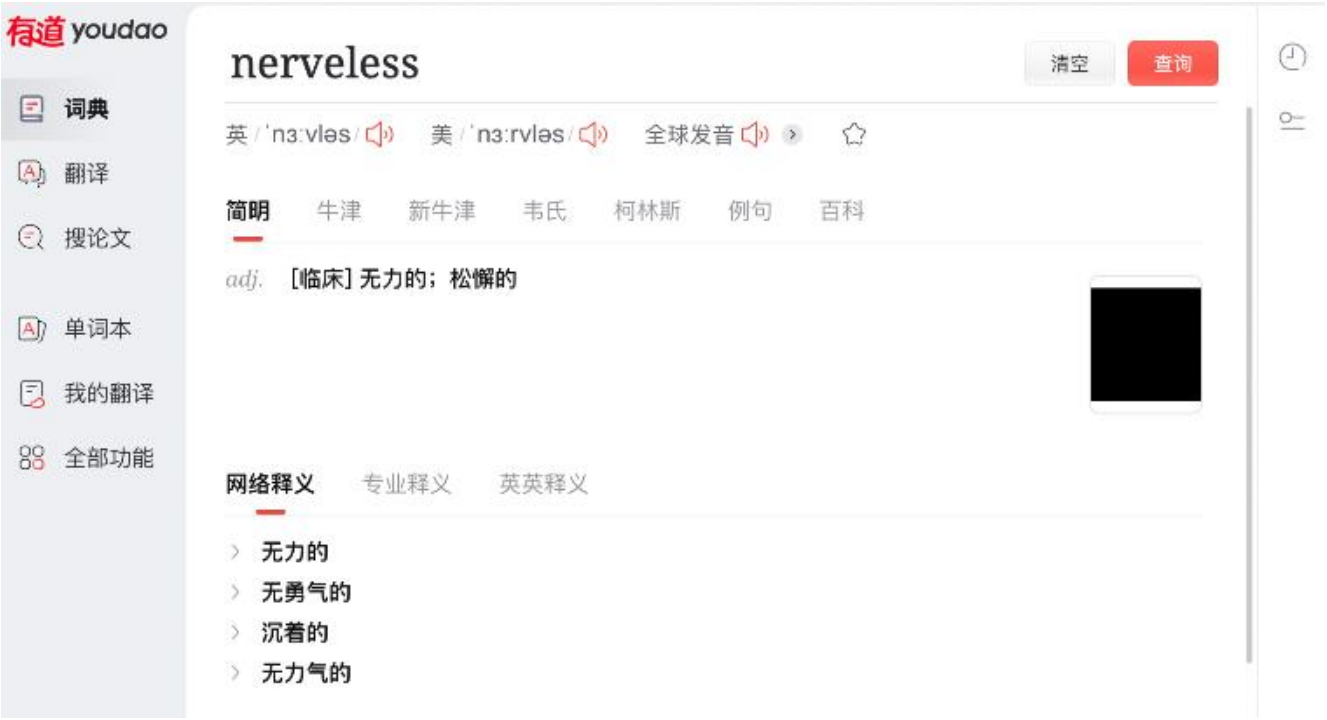

Figure 10. Screenshot of the concise dictionary's entry *nerveless*

Because of this, 34% of participants used the meaning 松懈的(song xie de, lax) or 无力的(wu li de, powerless) — provided in the CD in their translation, and got 0 point for this sentences. The use of other negative words, influenced by the definition in CD, as 没勇气 (mei yong qi, lacking courage), 差 (cha, worse), and 弱 (ruo, lacking strength) were also regarded as completely wrong:

他是足球史上最**松懈的**冠军之一。

(ta shi zu qiu li shi shang zui **song xie de** guan jun zhi yi)

(He is one of the **laxest** champions in football history.)

他是足球历史上最**无力的**冠军之一。

(ta shi zu qiu li shi shang zui **wu li de** guan jun zhi yi)

(He is one of the **weakest** champions in football history.)

他是足球历史上最**弱**的冠军之一。

(ta shi zu qiu li shi shang zui **ruo** de guan jun zhi yi)

(He is one of the **weakest** champions in football history.)

## 6.4. Confusing Definition

The understanding of the word *slosh* reveals another problem. Those who consulted the CD scored only 0.4 on average. In the Categorical summary analysis, 60% of those used the CD got 0 point, and less than 30% grasped the meaning of *slosh*. Participants' introspective questions reveals possible reasons. Some pointed out:

1. The expression of its verb is hard to understand.
2. I had to use the context to paraphrase 泥中荡 (ni zhong dang, swing in the mud) into "wade into the mud".
3. The Chinese phrase 在泥中荡(swing in the mud) mislead me, making me misunderstand it as something like "stir".

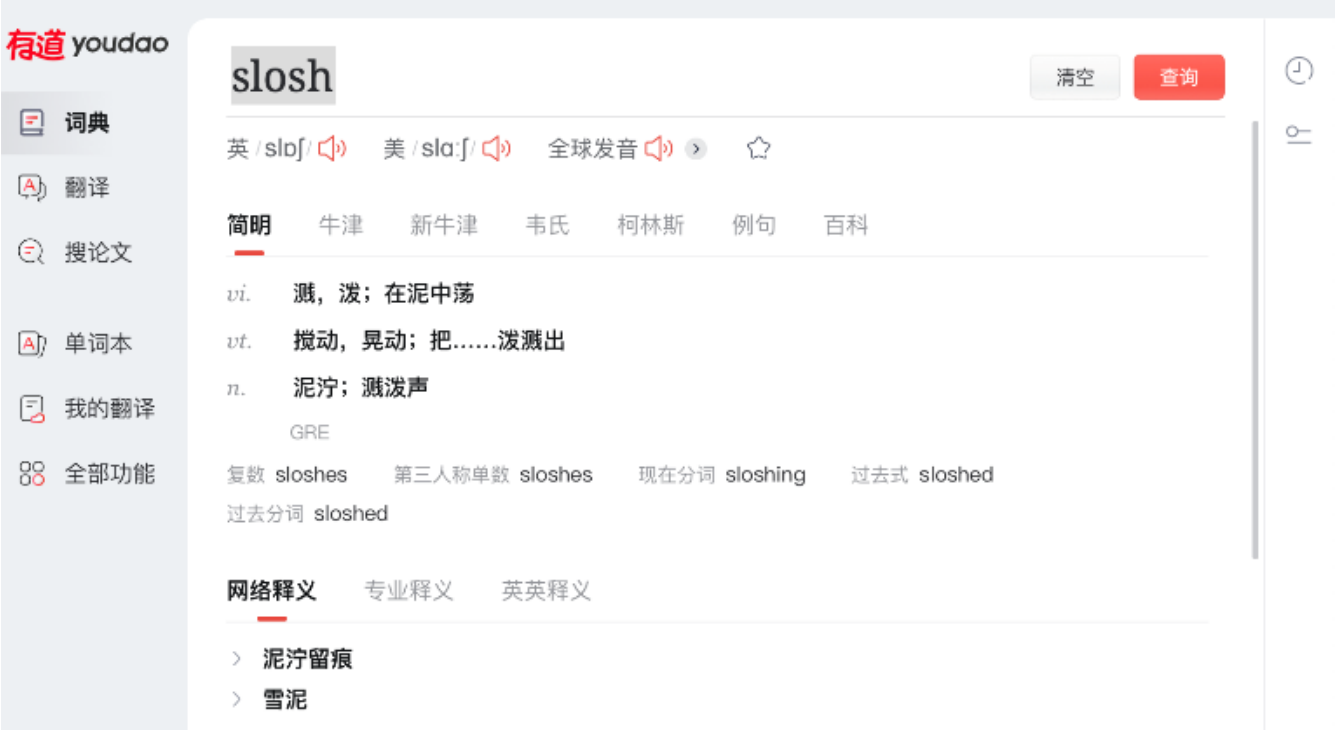

Figure 11. Screenshot of the concise dictionary's entry *slosh*

Indeed, 在泥中荡(swing in the mud) is not an idiomatic Chinese phrase, and it's hard to understand. The source text and reference answer are as follows:

The two girls joined arms and **sloshed** through the mud together.

这两个女孩挽着胳膊，一起**趟过**了泥浆。

zhe liangge nühai wan zhe gebo ， yiqi **tang guo** le nijiang.

Some participants used awkward expressions directly:

这两个女孩挽着手臂在泥中**荡着**。

zhe liang ge nü hai wan zhe shou bi zai ni zhong **dang zhe.**）

(The two girls **swang** in the slurry with arms entwined.）

这两个女孩挽着胳膊在泥浆中**荡漾**。

(zhe liang ge nü hai wan zhe ge bo zai ni jiang zhong **dang yang**.)

(The two girls **rippled** in the slurry with arms entwined.)

And some of them sought help from other definitions. For example, the second sense in CD, 撒出, 溅出(sa chu, jian chu), meaning "to move around making a lot of noise or coming out over the edge of sth":

两个女孩手臂交织着**泼洒**泥浆。

(liang ge nü hai shou bi jiao zhi zhe **po sa** ni jiang.)

(The two girls **splashed** the slurry with arms entwined.)

两个女孩相互缠绕手臂来**搅动**泥浆。

(liang ge nü hai xiang hu chan rao shou bi lai jiao dong ni jiang.)

(The two girls **stirred** the slurry with arms entwined.)

In the ADs, *slosh* is translated as（在水或泥里）扑哧扑哧地走 (zai shui huo ni li pu chi pu chi de zou), which means "to walk noisily in water or mud". This translation is evidently more comprehensible for users. Students used the ADs got higher marks with an increased score of 0.64.

## 6.5. Summary of the experiment

After finishing the translation experiment, some participants retrospected their dissatisfaction of the CD:

1. It provides far too less definitions.
2. For the polysemies, Youdao sometimes lacks one of the deifinitions.
3. The CD alone could not express connotational meanings through Chinese equivalents. There are many omissions in meanings.
4. I believe there are problems in Youdao Dictionary because it relies on machine translation, resulting in discrepancies between definitions and actual contextual meanings. Another issue is the significant difference between the definitions in the CD and the ADs.

The findings show that problematic definitions in dictionaries could have adverse effect on students' comprehension of source texts. There was a high probability that students would be misguided by inaccurate definitions, resulting in a completely incorrect interpretation. The non-idiomatic definitions and inadequate explanations might necessitate students to speculate the potential meanings. In this way, students had to either resort to other definitions that lacked contextual relevance, or deduce the meaning based on the limited information from the unidiomatic or deficient definitions. Still, in the two circumstances, the success rate of understanding implied meaning was considerably low.

# 7. Habits of E-dictionary use

Another aspect worth discussing in this project is the dictionary use habits of students. We assumed that relying on authoritative dictionaries should lead to better performance in translation, but it was not the case. Table 4 with the translation scores shows that, although students consulted the ADs, their average scores in rendering *dearly*, *slosh*, and *nostrum* were low compared to other words. To explore the reason behind this, we further analyzed the categorical summary analysis results of these three words and discovered a significant difference in students' reliance on CD and ADs.

Table 6 displays the results of the introspective question regarding how participants determine the meaning of *dearly*, indicating that students tend to rely too heavily on CD and show reluctance in consulting ADs. 65% of the participants claimed that their main understanding was from the CD. In contrast, only 35 out of 100 reconsulted the words in ADs for better understanding. This means more than half of the students are credulous with the word meanings in CD and thus did not think it necessary to visit ADs for better results. However, in this case, 94.3% of students who relyied on ADs gained full marks, but 61.5% of participants whose main source was CD completely misunderstood the word's meaning and got 0 points.

| | Number of students | Percentage |
|---|---|---|
| CD_false（0） | 40 | **61.5%** |
| CD_partially correct（1） | 2 | 3.0% |
| CD_correct（2） | 23 | 35.4% |
| **CD _total** | **65** | |
| ADs_ false（0） | 0 | 0 |
| ADs_ partially correct（1） | 2 | 5.7% |
| ADs_ correct（2） | 33 | **94.3%** |
| **ADs _total** | **35** | |
| Total | 100 | |

Table 6. Categorical summary analysis of *dearly*

While we further analyzed the translation answers of all participants for *dearly*, it was found that 24 % of the assumed answers generated from using ADs were identical to the answers in the CD. These were all wrong answers with 0 marks, thus dragging down the average score of using the AD, as shown in Table 4. Due to the sequence design of the translation experiment, the question based on the use of CD was presented before the one using ADs, meaning that the participants might have directly copied their previous answers to the latter translations. This assumption was confirmed by retrieving the screen recordings of the process. Therefore, it could be concluded that the students had a trust in the CD, and thus skipped the use of ADs.

Upon reviewing the AD translation answers of the word *slosh*, we discovered that 32% of translations adopted the synonyms of the first and second senses in ADs, resulting in misunderstandings of the word *slosh* in the context. After retrieving the screen recordings, the assumption was confirmed that these students tended to hover their mouse pointers over the first and second senses for an extended period without noticing the third sense, which was the correct meaning. This finding supports the previous study conducted by Nesi & Haill (2002) that dictionary users are prone to the first definition given for a polysemous word, even when it is not appropriate in the context.

To conclude, our retrospective section reveals two habits of dictionary users which may lead to negative learning effects: over-reliance on the inaccurate online dictionary and a tendency to focus solely on the head entries in dictionaries. Therefore, we recommend that students receive more instruction on dictionary skills to empower them to use authoritative dictionaries more effectively, overcome learning barriers caused by poor dictionary habits, and enhance their overall language learning experience.

## 8. Conclusion

Taking a combined research method of a questionnaire, a consultation experiment and translation retrospection, this research confirmed that the most popular Chinese-English bilingual online dictionary, Youdao, was predominantly used by its users with major reference to its Concise section. However, the presence of erroneous, insufficient, and incomprehensible definitions in the CD could adversely affect students' comprehension of words and sentences. It is recommended that students develop good dictionary habits by consulting authoritative dictionaries more frequently and carefully read each sense to find the most suitable meaning. For online dictionary developers, it is crucial to rectify all problematic definitions and provide high-quality assistance for word lookups.

The current study primarily focused on the influence of problematic definitions. In future studies, we will delve into their impact on paragraph comprehension and vocabulary acquisition. Furthermore, exploring the extent to which the quality of online dictionary definitions affects students' performance in productive tasks remains an area for further investigation. From a broader technical perspective, there is a need to explore the reasons for Youdao Dictionary's interpretation errors and put forward improvement plans. It is an inevitable trend for students to use e-learning tools to assist language learning in today's artificial intelligence-driven society, but ensuring the quality and effectiveness of these reference tools remains a critical prerequisite.

## Acknowledgements

This project was sponsored by the Undergraduate Research Award (URA) funding of the Chinese University of Hong Kong, Shenzhen.

## Appendix 1: translation experiment design

The bold words are the targeted tricky ones (which are not bold or highlighted in the task sheet):

1. Freedom to worship our religion has been bought **dearly**.
2. *The discussion is getting more and more* ***furious****.*
3. *He admitted that this problem is a main* ***irritant****.*
4. *They launch a* ***vigorous*** *movement to oppose corruption.*
5. *The two girls* ***sloshed*** *through the slurry with arms entwined.*
6. *Almost everyone in the audience felt this comedy was a bit* ***raunchy****.*
7. *Although my sister is not a doctor, she thinks she can cure any illness and is quick to suggest a* ***nostrum*** *to her friends.*
8. *I am a bit* ***dicky*** *today.*
9. *It is not just physicists who have a* ***fetish*** *for expensive machines.*
10. *He is one of the most* ***nerveless*** *champions in football history.*

## Appendix 2: questionnaire design

1. Basic Background Information of the Participants
    a. What's your major?
    b. What grade are you in?
    c. What score did you get on NMET (National Matriculation English Test)?
2. Use Habits of the Youdao Dictionary
    a. How often do you use the Youdao Dictionary? (Factor:FrequencyUsage)
    b. In what English learning situations do you usually use Youdao Dictionary?
        i. Listening (Factor: ListeningDifficulty)
        ii. Speaking (Factor: SpeakingDifficulty)
        iii. Reading (Factor: ReadingDifficulty)
        iv. Writing (Factor: WritingDifficulty)
        v. Translating (Factor: TranslationDifficulty)
    c. Which dictionary do you most often use when searching for new words you come across in English listening and reading? The concise dictionary or the authoritative dictionaries?
    d. (Factor: ListeningReading)
    e. Which dictionary do you most often use when searching for new words you come across in English speaking, writing, and translating? The concise dictionary or the authoritative dictionaries? (Factor: WritingTranslating)
3. Attitudes Toward Youdao
    a. What's your satisfaction level with the definition provided by the concise dictionary? (Factor: SatisfactionLevel)
    b. To what extent do you agree with the statement:
        i. Youdao Dictionary can help you solve vocabulary problems in English learning. (Factor: VocabularyAid)
        ii. Youdao dictionary can help me understand English readings correctly. (Factor: ReadingAid)
        iii. Using Youdao Dictionary can fully meet your need to find words (Factor: WordLookUp)